\documentclass[runningheads]{llncs}
\usepackage{graphicx}

\usepackage{tikz}
\usepackage{comment} 
\usepackage{amsmath,amssymb} 

\usepackage{color}
\usepackage{booktabs}
\usepackage{array}
\usepackage{multirow}
\usepackage{color}
\usepackage{hyperref}
\hypersetup{
    colorlinks=true,
    linkcolor=magenta,
    filecolor=magenta,      
    urlcolor=magenta,
}

\usepackage{enumitem}
\usepackage{wrapfig}
\usepackage{float}

\newfloat{figtab}{htb}{fgtb}
\makeatletter
  \newcommand\figcaption{\def\@captype{figure}\caption}
  \newcommand\tabcaption{\def\@captype{table}\caption}
\makeatother

\def\ie{\textit{i.e.}}
\def\eg{\textit{e.g.}}
\def\wrt{\textit{w.r.t. }}
\def\vs{\textit{v.s. }}

\begin{document}
\pagestyle{headings}
\mainmatter
\def\ECCVSubNumber{100}  

\title{Towards Real-Time Multi-Object Tracking} 


\titlerunning{Towards Real-Time MOT}
%
\author{Zhongdao Wang\inst{1} \and
Liang Zheng\inst{2} \and
Yixuan Liu\inst{1} \and
Yali Li\inst{1} \and
Shengjin Wang\inst{1} }
\authorrunning{Z. Wang et al.}
%
\institute{Department of Electronic Engineering, Tsinghua University
\email{wcd17@mails.tsinghua.edu.cn, \{liyali13, wgsgj\}@tsinghua.edu.cn}\\ \and
Australian National University\\
\email{liang.zheng@anu.edu.au}}
\maketitle

\begin{abstract}
Modern multiple object tracking (MOT) systems usually follow the \emph{tracking-by-detection} paradigm. It has 1) a detection model for target localization and 2) an appearance embedding model for data association. Having the two models separately executed might lead to efficiency problems, as the running time is simply a sum of the two steps without investigating potential structures that can be shared between them. Existing research efforts on real-time MOT usually focus on the association step, so they are essentially real-time association methods but not real-time MOT system. 
In this paper, we propose an MOT system that allows target detection and appearance embedding to be learned in a shared model. 
Specifically, we incorporate the appearance embedding model into a single-shot detector, such that the model can simultaneously output detections and the corresponding embeddings.
We further propose a simple and fast association method that works in conjunction with the joint model. In both components the computation cost is significantly reduced compared with former MOT systems, resulting in a neat and fast baseline for future follow-ups on real-time MOT algorithm design.
To our knowledge, this work reports the first (near) real-time MOT system, with a running speed of 22 to 40 FPS depending on the input resolution. Meanwhile, its tracking accuracy is comparable to the state-of-the-art trackers embodying separate detection and embedding (SDE) learning ($64.4\%$ MOTA \vs $66.1\%$ MOTA on MOT-16 challenge). Code and models are available at \url{https://github.com/Zhongdao/Towards-Realtime-MOT}. 

\end{abstract}

\section{Introduction}
Multiple object tracking (MOT), which aims at predicting trajectories of multiple targets in video sequences, underpins critical application significance ranging from autonomous driving to smart video analysis. 

The dominant strategy to this problem, \ie,  \emph{tracking-by-detection} ~\cite{mot16,poi,nomt} paradigm, breaks MOT down to two steps: 1) the detection step, in which targets in single video frames are localized; and 2) the association step, where detected targets are assigned and connected to existing trajectories. 
It means the system requires at least two compute-intensive components: a detector and an embedding (re-ID) model. We term those methods as the Separate Detection and Embedding (SDE) methods for convenience. The overall inference time, therefore, is roughly the summation of the two components, and will increase as the target number increases. {The characteristics  of SDE methods bring critical challenges in building a real-time MOT system, an essential demand in practice.}
\begin{figure}[t]
    \centering
    \includegraphics[width=\linewidth]{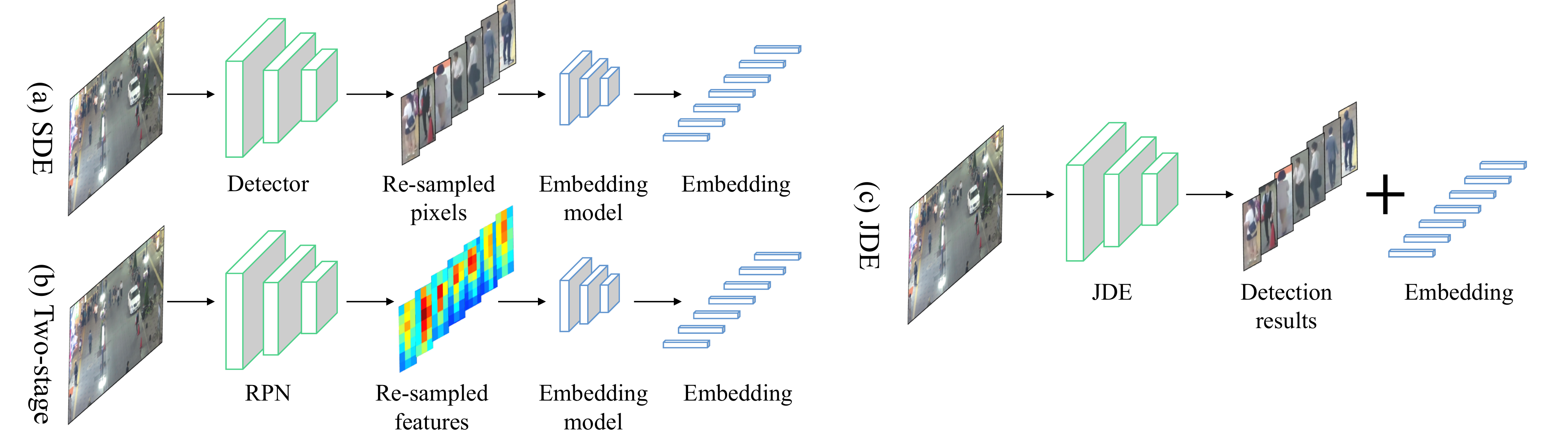}
    \caption{Comparison between (a) the Separate Detection and Embedding (SDE) model, (b) the two-stage model and (c) the proposed Joint Detection and Embedding (JDE).}
    \label{fig:intro}
\end{figure}
    
In order to save computation, a feasible idea is to integrate the detector and the embedding model into a single network. The two tasks thus can share the same set of low-level features, and re-computation is avoided. One choice for joint detector and embedding learning is to adopt the Faster R-CNN framework~\cite{faster}, a type of two-stage detectors. Specifically, the first stage, the region proposal network (RPN), remains the same with Faster R-CNN and outputs detected bounding boxes; the second stage, Fast R-CNN~\cite{fast}, can be converted to an embedding model by replacing the classification supervision with the metric learning supervision~\cite{personsearch,MOTS}. In spite of saving some computation, this method is still limited in speed due to its two-stage design and usually runs at fewer than 10 frames per second (FPS), far from real-time. Moreover, the runtime of the second stage also increases as target number increases like SDE methods.

This paper is dedicated to the improving efficiency of an MOT system. We introduce an early attempt that Jointly learns the Detector and Embedding model (JDE) in a \emph{single-shot} deep network. In other words, the proposed JDE employs a single network to \emph{simultaneously} output detection results and the corresponding appearance embeddings of the detected boxes. In comparison, SDE methods and two-stage methods are characterized by re-sampled pixels (bounding boxes) and feature maps, respectively. Both the bounding boxes and feature maps are fed into a separate re-ID model for appearance feature extraction. Figure~\ref{fig:intro} briefly illustrates the difference between the SDE methods, the two-stage methods and the proposed JDE. Our method is near real-time while being  almost as accurate as the SDE methods. For example, we obtain a running time of 20.2 FPS with MOTA=$64.4\%$ on the MOT-16 test set. In comparison, Faster R-CNN + QAN embedding~\cite{poi} only runs at $<$6 FPS with MOTA=$66.1\%$ on the MOT-16 test set.

To build a joint learning framework with high efficiency and accuracy, we explore and deliberately design the following fundamental aspects:  training data, network architecture,  learning objectives, optimization strategies, and validation metrics.
First, we collect six publicly available datasets on pedestrian detection and person search to form a unified large-scale multi-label dataset. In this unified dataset, all the pedestrian bounding boxes are labeled, and a portion of the pedestrian identities are labeled. Second, we choose the Feature Pyramid Network (FPN)~\cite{fpn} as our base architecture and discuss with which type of loss functions the network learns the best embeddings. Then, we model the training process as a multi-task learning problem with anchor classification, box regression, and embedding learning. To balance the importance of each individual task, we employ task-dependent uncertainty~\cite{uncertainty} to dynamically weight the heterogenous losses.
A simple and fast association algorithm is proposed to further improve efficiency. 
Finally, we employ the following evaluation metrics. The average precision (AP) is employed to evaluate the performance of the detector. The retrieval metric True Accept Rate (TAR) at certain False Alarm Rate (FAR) is adopted to evaluate the quality of the embedding. The overall MOT accuracy is evaluated by the CLEAR metrics~\cite{CLEAR}, especially the MOTA score. This paper also provides  new settings and baselines for  joint detection and embedding learning, which we believe will facilitate research towards real-time MOT. 

The contributions of our work are summarized as follows, 

\begin{itemize}
    \item We introduce JDE, a single-shot framework for joint detection and embedding learning. It runs in (near) real-time and is comparably accurate to the separate detection + embedding (SDE) state-of-the-art methods.
    \item We conduct thorough analysis and experiments on how to build such a joint learning framework from multiple aspects including training data, network architecture, learning objectives and  optimization strategy.
    \item Experiments with the same training data show the JDE performs as well as  a range of strong SDE model combinations and achieves the fastest speed.
    \item Experiments on  MOT-16 demonstrate the advantage of our method over state-of-the-art MOT systems considering the amount of training data, accuracy and speed. 
\end{itemize}




\section{Related Work}

Recent progresses on multiple object tracking can be primarily categorized into the following aspects: 
\begin{enumerate}
    \item[1)] Ones that model the association problem as certain form of optimization problem on graphs~\cite{graph1,graph2,graph3}.
    \item[2)] Ones that make efforts to model the association process by an end-to-end neural network~\cite{e2e1,e2e2}.
    \item[3)] Ones that seek novel tracking paradigm other than tracking-by-detection~\cite{tracktor}.
\end{enumerate}

Among them, the first two categories have been the prevailing solution to MOT in the past decade. In these methods, detection results and appearance embeddings are given as input, and the only problem to be solved is data association. A standard formulation is using a graph, where nodes represent a detected targets, and edges indicate the possibility of linkages among nodes. Data association thus can be solved by minimizing some fixed~\cite{mot1,mot2,mot3} or learned~\cite{mot4} cost, or by more complex optimization such as multi-cuts~\cite{multicut} and minimum cliques~\cite{mot5}. Some 
recent works attempt to model the association problem using graph networks~\cite{gnnmot1,gnnmot2}, so that end-to-end association can be achieved. Graph-based association shows good tracking accuracy especially in hard cases such as large occlusions, but their efficiency is always a problem.  Although some methods~\cite{nomt} claim to be able to attain real-time speed, the runtime of the detector is excluded, such that the overall system still has some distance from the claim. In contrast, in this work, we consider the runtime of \emph{the entire MOT system} rather than the association step only. Achieving efficiency on the entire system is more practically significant. 

The third category attempts to explore novel MOT paradigms, for instance, incorporating single object trackers into the detector by predicting the spatial offsets~\cite{tracktor}. These methods are appealing owning to their simplicity, but tracking accuracy is not satisfying unless an additional embedding model is introduced. As such, the trade-off between performance and speed still needs improvement.

 The spirit of our approach, that learning auxiliary associative embeddings simultaneously with the main task, also shows good performance in many other vision tasks, such as person search~\cite{personsearch}, human pose estimation~\cite{ae}, and point-based object detection~\cite{cornernet}.



\section{Joint Learning of Detection and Embedding}
\subsection{Problem Settings}
\label{sec:problemdef}
The objective of JDE is to simultaneously output the location and appearance embeddings of targets in a single forward pass. Formally, suppose we have a training dataset $\{\mathbf{I}, \mathbf{B}, \mathbf{y}\}_{i=1}^N$. Here, $\mathbf{I} \in \mathbb{R}^{c\times h \times w}$ indicates an image frame, and $\mathbf{B} \in \mathbb{R}^{k\times 4}$ represents the bounding box annotations for the $k$ targets in this frame. $\mathbf{y} \in \mathbb{Z}^{k}$ denotes the partially annotated identity labels, where $-1$ indicates targets without an identity label. JDE aims to output predicted bounding boxes $\hat{\mathbf{B}} \in \mathbb{R}^{\hat{k}\times 4}$ and appearance embeddings $\hat{\mathbf{F}} \in  \mathbb{R}^{\hat{k} \times D}$, where $D$ is the dimension of the embedding. The following objectives should be satisfied.
\begin{itemize}
    \item $\mathbf{B^{*}}$ is as close to $\mathbf{B}$ as possible.
    \item Given a distance metric $d(\cdot)$, $\forall (k_t, k_{t+\Delta t}, k_{t+\Delta t}^{\prime})$ that satisfy $\mathbf{y}_{k_{t+\Delta t}} = \mathbf{y}_{k_t}$ and   $\mathbf{y}_{k_{t+\Delta t}^{\prime}} \ne \mathbf{y}_{k_t}$, we have $d(f_{k_t},f_{k_{t+\Delta t}}) < d(f_{k_t},f_{k_{t+\Delta t}^{\prime}})$, where $f_{k_t}$ is a row vector from $ \hat{\mathbf{F}}_t$ and $f_{k_{t+\Delta t}}, f_{k_{t+\Delta t}^{\prime}}$ are row vectors from $ \hat{\mathbf{F}}_{t+\Delta t}$, \ie, embeddings of targets in frame $t$ and ${t+\Delta t}$, respectively, 
\end{itemize}

The first objective requires the model to detect targets accurately. The second objective requires the appearance embedding to have the following property. The distance between observations of the same identity in consecutive frames should be smaller than the distance between different identities. The distance metric $d(\cdot)$ can be the Euclidean distance or the cosine distance. Technically, if the two objectives are both satisfied, even a simple association strategy, \emph{e.g.}, the Hungarian algorithm, would produce good tracking results.

\begin{figure}[t]
    \centering
    \includegraphics[width=\linewidth]{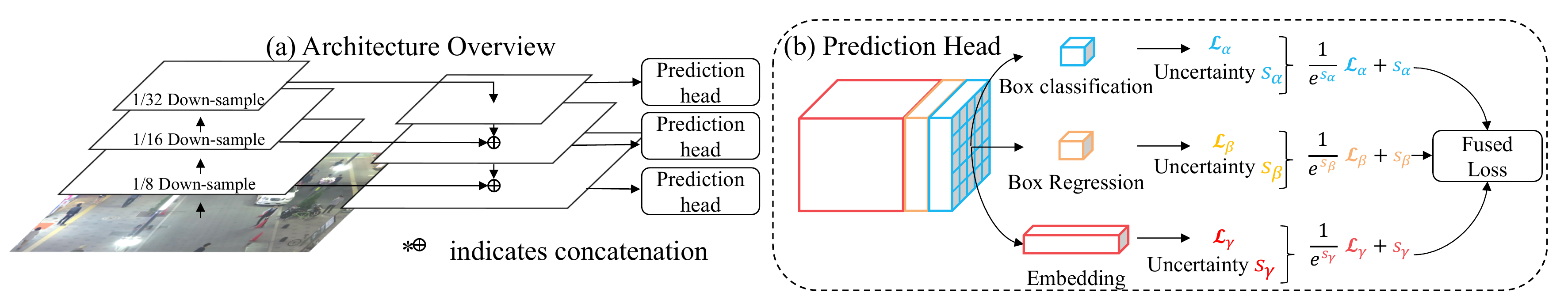}
    \caption{Illustration of (a) the network architecture and (b) the prediction head. Prediction heads are added upon multiple FPN scales. In each prediction head the learning of JDE is modeled as a multi-task learning problem. We automatically weight the heterogeneous losses by learning a set of auxiliary parameters, \ie, the task-dependent uncertainty.}
    \label{fig:architecture}
\end{figure}
\subsection{Architecture Overview}
We employ the architecture of Feature Pyramid Network (FPN) \cite{fpn}. FPN makes predictions from multiple scales, thus bringing improvement in pedestrian detection where the scale of targets varies a lot. Figure~\ref{fig:architecture} briefly shows the neural architecture used in JDE. 
An input video frame first undergoes a forward pass through a backbone network to obtain feature maps at three scales, namely, scales with $\frac{1}{32}$, $\frac{1}{16}$ and $\frac{1}{8}$ down-sampling rate, respectively. Then, the feature map with the smallest size (also the semantically strongest features) is up-sampled and fused with the feature map from the second smallest scale by skip connection, and the same goes for the other scales. Finally, prediction heads are added upon fused feature maps at all the three scales. A prediction head consists of several stacked convolutional layers and outputs a dense prediction map of size $(6A+D)\times H\times W$, where $A$ is the number of anchor templates assigned to this scale, and $D$ is the dimension of the embedding. The dense prediction map is divided into three parts (tasks):
\begin{itemize}
    \item[1)] the box classification results of size $2A\times H\times W$;
    \item[2)] the box regression coefficients of size $4A\times H\times W$; 
    \item[3)] the dense embedding map of size $D\times H\times W$.
\end{itemize}

In the following sections, we will detail how these tasks are trained.

\subsection{Learning to Detect}
In general the detection branch is similar to the standard RPN~\cite{faster}, but with two modifications. First, we redesign the anchors in terms of numbers, scales, and aspect ratios to be able to adapt to the targets, \ie, pedestrian in our case. Based on the common prior, all anchors are set to an aspect ratio of $1:3$. The number of anchor templates is set to $12$ such that $A=4$ for each scale, and the scales (widths) of anchors range from $11 \approx 8\times2^{1/2}$ to $512=8\times2^{12/2}$.
Second, we note that it is important to select proper values for the dual thresholds used for  foreground/background assignment. By visualization we determine that an IOU$>$0.5 \wrt the ground truth approximately ensures a foreground, which is consistent with the common setting in generic object detection. On the other hand, those boxes that have an IOU$<$0.4 \wrt the ground truth should be regarded as background in our case rather than $0.3$ used in generic scenarios.
Our preliminary experiment indicates that these thresholds effectively suppress false alarms, which usually happens under heavy occlusions. 

The learning objective of detection has two loss functions, namely the foreground/background classification loss $\mathcal{L}_\alpha$, and the bounding box regression loss $\mathcal{L}_\beta$.  $\mathcal{L}_\alpha$ is formulated as a cross-entropy loss and $\mathcal{L}_\beta$ as a smooth-L1 loss. The regression targets are encoded in the same manner as \cite{faster}.

\subsection{Learning Appearance Embeddings}
The second objective is a \emph{metric learning} problem, \emph{i.e.,} learning a embedding space where instances of the same identity are close to each other while instances of different identities are far apart.
To achieve this goal, an effective solution is to use the triplet loss~\cite{triplet}. The triplet loss has also been used in previous MOT works \cite{MOTS}. Formally, we use triplet loss
$\mathcal{L}_{triplet} = \max (0, f^{\top}f^{-} - f^{\top}f^{+})$,
where $f^{\top}$ is an instance in a mini-batch selected as an anchor, $f^{+}$ represents a positive sample \wrt $f^{\top}$, and $f^{-}$ is a negative sample. The margin term is neglected for convenience.
This naive formulation of the triplet loss has several challenges. The first is the huge sampling space in the training set. 
In this work we address this problem by looking at a mini-batch and mining all the negative samples and the hardest positive sample in this mini-batch, such that,
\begin{equation}
   \mathcal{L}_{triplet} = \sum_i \max \big(0, f^{\top}f^{-}_i - f^{\top}f^{+}\big),
\end{equation}
where $f^{+}$ is the hardest positive sample in a mini-batch.

The second challenge is that training with the triplet loss can be unstable and the convergence might be slow. To stabilize the training process and speed up convergence, it is proposed in~\cite{npair} to optimize over a smooth upper bound of the triplet loss,
\begin{equation}
    \mathcal{L}_{upper} = \log \Big(1+ \sum_i   \exp \big(f^{\top}f^{-}_i - f^{\top}f^{+}\big)\Big).
\end{equation}

Note that this smooth upper bound of triplet loss can be also written as,
\begin{equation}
    \mathcal{L}_{upper} = - \log \frac{\exp(f^{\top}f^+) }{\exp(f^{\top}f^+) + \sum_i \exp(f^{\top}f^{-}_i) }.
\end{equation}
It is similar to the formulation of the cross-entropy loss,
\begin{equation}
    \mathcal{L}_{CE} = - \log \frac{\exp(f^{\top}g^+) }{\exp(f^{\top}g^+) + \sum_i \exp(f^{\top}g^{-}_i) },
\end{equation}
where we denote the class-wise weight of the positive class (to which the anchor instance belongs) as $g^+$ and weights of negative classes as $g_i^-$. The major ditinctions between $\mathcal{L}_{upper}$ and $\mathcal{L}_{CE}$ are two-fold. First, the cross-entropy loss employs learnable class-wise weights as proxies of class instances rather than using the embeddings of instances directly. Second, all the negative classes participate in the loss computation in $\mathcal{L}_{CE}$ such that the anchor instance is pulled away from all the negative classes in the embedding space. In contrast, in $\mathcal{L}_{upper}$, the anchor instance is only pulled away from the sampled negative instances.

In light of the above analysis, we speculate the performance of the three losses under our case should be $\mathcal{L}_{CE} > \mathcal{L}_{upper} > \mathcal{L}_{triplet}$. Experimental result in the experiment section confirms this. As such, we select the cross-entropy loss as the objective for embedding learning (hereinafter referred to as $\mathcal{L}_{\gamma}$).

Specifically, if an anchor box is labeled as the foreground, the corresponding embedding vector is extracted from the dense embedding map. Extracted embeddings are fed into a \emph{shared} fully-connected layer to output the class-wise logits, and then the  cross-entropy loss is applied upon the logits. In this manner, embeddings from multiple scales shares the same space, and association across scales is feasible. Embeddings with label $-1$, \ie, foregrounds with box annotations but without identity annotations, are ignored when computing the embedding loss. 

\subsection{Automatic Loss Balancing}
\label{sec:lossbalancing}
The learning objective of each prediction head in JDE can be modeled as a multi-task learning problem. The joint objective can be written as a weighted linear sum of losses from every scale and every component,
\begin{equation}
    \mathcal{L}_{total} = \sum_{i}^M \sum_{j=\alpha, \beta, \gamma} w_{j}^{i} \mathcal{L}_{j}^{i},
\end{equation}
where $M$ is the number of prediction heads and $w_j^i, i=1,...,M, j=\alpha, \beta, \gamma$ are loss weights. 
A simple way to determine loss weights are described below.
\begin{enumerate}
    \item Let $w_\alpha^i = w_\beta^i$, as suggested in~\cite{faster}
    \item Let $w_{\alpha/\gamma/\beta}^1 = ... = w_{\alpha/\gamma/\beta}^M$.
    \item Search for the remaining two independent loss weights for the best performance.
\end{enumerate}
Searching loss weights with this strategy can yield decent results within several attempts. However, the reduction of searching space also brings strong restrictions on the loss weights, such that the resulting loss weights might be far from optimal. 
Instead, we adopt an automatic learning scheme for loss weights proposed in~\cite{uncertainty} by using the concept of task-independent uncertainty. Formally, the learning objective with automatic loss balancing is written as,
\begin{equation}
\begin{aligned}
        \mathcal{L}_{total} = \sum_{i}^M \sum_{j=\alpha, \beta, \gamma}\frac{1}{2}\left( \frac{1}{e ^{s_{j}^i}} \mathcal{L}_{j}^{i} 
         +  s_{j}^i \right),
\end{aligned}
\end{equation}
where $s_j^{i}$ is the task-dependent uncertainty for each individual loss and is modeled as learnable parameters. We refer readers to \cite{uncertainty} for more detailed derivation and discussion.

\subsection{Online Association}
\label{sec:association}

\begin{wraptable}{r}{0.6\linewidth}

    \centering
    \begin{tabular}{l | c | c  c  c }
    \toprule
    Method & Density & FPS & MOTA & IDF-1 \\
    \hline
    SORT~\cite{sort} & low & 44.1 & 66.9 & 55.8\\
    ours & low & \textbf{46.2} & \textbf{67.5} & \textbf{67.6}\\
    
    \hline
    SORT~\cite{sort} & high & 26.4 & 35.0 & 32.4\\
    ours & high & \textbf{33.9} & \textbf{35.4} & \textbf{35.5}\\
    \bottomrule
    
    \end{tabular}
    \caption{Comparison between our association method and SORT. Inputs are the same. }
    \label{tab:sort}
    
\end{wraptable}

Although the association algorithm is not the focus of this work, here we introduce a simple and fast online association strategy to work in conjunction with JDE. Specifically, a tracklet is described with an appearance state ${e}_i$ and a motion state ${m_i} = (x,y,\gamma, h, \dot{x}, \dot{y}, \dot{\gamma}, \dot{h})$, where $x,y$ indicate the bounding box center position, $h$ indicates the bounding box height and $\gamma$ indicates the aspect ratio, and $\dot{x}$ indicates the velocity along $x$ direction. The tracklet appearance $e_i$ is initialized with the appearance embedding of the first observation $f_i^0$. We maintain a tracklet pool containing all the reference tracklets that observations are probable to be associated with. For an incoming frame, we compute the pair-wise motion affinity matrix $A_m$ and appearance affinity matrix $A_e$ between all the observations and the traklets from the pool. The appearance affinity is computed using cosine similarity, and the motion affinity is computed using Mahalanobis distance. Then we solve the linear assignment problem by Hungarian algorithm with cost matrix $C=\lambda A_e + (1-\lambda) A_m$. The motion state $m_i$ of all matched tracklets are  updated by the Kalman filter, and the appearance state $e_i$ is updated by
\begin{equation}
    e_i^t = \alpha e^{t-1}_i + (1-\alpha) f_i^t
\end{equation}
Where $f_i^t$ is the appearance embedding of the current matched observation, $\alpha=0.9$ is a momentum term. Finally observations that are not assigned to any tracklets are initialized as new tracklets if they consecutively appear in 2 frames. A tracklet is terminated if it is not updated in the most current $30$ frames.

Note this association method is simpler than the cascade matching strategy proposed in SORT~\cite{sort}, since we only apply association once for one frame and resort to a buffer pool to deal with those shortly lost tracklets. Moreover, we also implement a vectorized version of the Kalman filter and find it critical for high FPS, especially when the model is already fast. A comparison between SORT and our association method, based on the same JDE model, is shown in Table~\ref{tab:sort}. We use MOT-15~\cite{mot16} for testing the low density scenario and CVPR-19-01~\cite{mot19} for high density. It can be observed that our method outperforms SORT in both accuracy and speed, especially under the high-density case.


\section{Experiments}
\subsection{Datasets and Evaluation Metrics}

\begin{wraptable}{R}{0.6\linewidth}

    \centering
    \begin{tabular}{l|ccccccc}
    \toprule
     Dataset& ETH & CP & CT & M16 & CS & PRW & Total \\
     \midrule
     $\#$ img & 2K & 3K& 27K& 53K& 11K& 6K& 54K\\
     $\#$ box & 17K& 21K& 46K& 112K& 55K& 18K& 270K\\
     $\#$ ID & -& - & 0.6K & 0.5K& 7K& 0.5K& 8.7K\\
     \bottomrule
\end{tabular}
    \caption{Statistics of the joint training set.}
    \label{tab:dataset}
\end{wraptable}

Performing experiments on small datasets may lead to biased results and conclusions may not hold when applying the same algorithm to large-scale datasets. 
Therefore, we build a large-scale training set by putting together six publicly available  datasets on pedestrian detection, MOT and person search. These datasets can be categorized into two types: ones that only contain bounding box annotations, and ones that have both bounding box and identity annotations. The first category includes the ETH dataset~\cite{eth} and the CityPersons (CP) dataset~\cite{citypersons}. The second category includes the CalTech (CT) dataset~\cite{caltech}, MOT-16 (M16) dataset~\cite{mot16}, CUHK-SYSU (CS) dataset~\cite{personsearch} and PRW dataset~\cite{prw}. Training subsets of all these datasets are gathered to form the joint training set, and videos in the ETH dataset that overlap with the MOT-16 \texttt{test} set are excluded for fair evaluation. 
Table~\ref{tab:dataset} shows the statistics of the joint training set.

For validation/evaluation, three aspects of performance need to be evaluated: the detection accuracy, the discriminative ability of the embedding, and the tracking performance of the entire MOT system. To evaluate detection accuracy, we compute average precision (AP) at IOU threshold of $0.5$ over the Caltech validation set. To evaluate the appearance embedding, we extract embeddings of all ground truth boxes over the validation sets of the Caltech dataset, the CUHK-SYSU dataset and the PRW dataset, apply $1:N$ retrieval among these instances and report the true positive rate at false accept rate $0.1$ (TPR@FAR=0.1). To evaluate the tracking accuracy of the entire MOT system, we employ the CLEAR metric~\cite{CLEAR}, particularly the MOTA metric that aligns best with human perception. In validation, we use the MOT-15 \texttt{training} set with duplicated sequences with the training set removed. During testing, we use the MOT-16 \texttt{test} set to compare with existing methods.

\subsection{Implementation Details}
We employ DarkNet-53~\cite{yolov3} as the backbone network in JDE. The network is trained with standard SGD for 30 epochs. The learning rate is initialized as $10^{-2}$ and is decreased by 0.1 at the 15th and the 23th epoch. Several data augmentation techniques, such as random rotation, random scale and color jittering, are applied to reduce overfitting. Finally, the augmented images are adjusted to a fixed resolution. The input resolution is $1088\times608$ if not specified.

\begin{wraptable}{R}{0.6\linewidth}

\begin{tabular}{l|l|c|c|cc}
    \toprule
         Embed. & Weighting  & Det & Emb & \multicolumn{2}{c}{MOT} \\
         \cline{3-6}
         Loss & Strategy & AP$\uparrow$ & TPR$\uparrow$ & MOTA$\uparrow$ &%
         IDs$\downarrow$ \\
        \midrule
         $\mathcal{L}_{triplet}$& App.Opt &81.6 &42.2& 59.5 & 375   \\
         $\mathcal{L}_{upper}$& App.Opt &81.7 &44.3& 59.8 & 346   \\
         $\mathcal{L}_{CE}$& App.Opt &\underline{82.0} &88.2&\underline{64.3} &\underline{223}   \\
         $\mathcal{L}_{CE}$& Uniform &6.8 &\textbf{94.8}& 36.9& 366   \\
         $\mathcal{L}_{CE}$& MGDA-UB &8.3&\underline{93.5}&38.3 & 357   \\
         $\mathcal{L}_{CE}$& Loss.Norm &80.6 &82.1& 57.9 & 321   \\
         $\mathcal{L}_{CE}$& Uncertainty &\textbf{83.0} &90.4&\textbf{65.8} &\textbf{207}   \\
    \bottomrule
       
    \end{tabular}
    
    \caption{Comparing different embedding losses and loss weighting strategies. TPR is short for TPR@FAR=0.1 on the embedding validation set, and IDs means times of ID switches on the tracking validation set. $\downarrow$ means the smaller the better; $\uparrow$ means the larger the better. In each column, the best result is in \textbf{bold}, and the second best is \underline{underlined}. }
    \label{tab:ablation}
\end{wraptable}

\subsection{Experimental Results}
\textbf{Comparison of the three loss functions for appearance embedding learning.} 
We first compare the discriminative ability of appearance embeddings trained with the cross-entropy loss, the the triplet loss and its upper bound variant, described in the previous section. 
For models trained with $\mathcal{L}_{triplet}$ and $\mathcal{L}_{upper}$, $B/2$ pairs of temporal consecutive frames are sampled to form a mini-batch with size $B$. This ensures that there always exist positive samples. For models trained with $\mathcal{L}_{CE}$, images are randomly sampled to form a mini-batch. Table~\ref{tab:ablation} presents comparisons of the three loss functions. 

As expected, $\mathcal{L}_{CE}$ outperforms both $\mathcal{L}_{triplet}$ and $\mathcal{L}_{upper}$. Surprisingly, the performance gap is large (+46.0/+43.9 TAR@FAR=0.1). A possible reason for the large gap is that the cross-entropy loss requires the similarity between one instance and its positive class be higher than the similarities between this instance and \emph{all}  negative classes. This objective is more rigorous than the triplet loss family, which exerts constraints merely in a sampled mini-batch. Considering its effectiveness and simplicity, we adopt the cross-entropy loss in JDE.

\textbf{Comparison of different loss weighting strategies.}
The loss weighting strategy is crucial to learn good joint representation for JDE.
In this paper, three loss weighting strategies are implemented. The first is a loss normalization method (named ``Loss.Norm''), where the losses are weighted by the reciprocal of their moving average magnitude. The second is the ``MGDA-UB'' algorithm proposed in~\cite{MGDA} and the last is the weight-by-uncertainty strategy described in Section~\ref{sec:lossbalancing}.
Moreover, we have two baselines. The first trains all the tasks with identical loss weights, named as ``Uniform''. The second, referred to as ``App.Opt'', uses a set of approximate optimal loss weights by searching under the two-independent-variable assumption as described in Section~\ref{sec:lossbalancing}. Table~\ref{tab:ablation} summarizes the comparisons of these strategies. Two observations are made. 

First, the Uniform baseline produces poor detection results, and thus the tracking accuracy is not good. This is because the scale of the embedding loss is much larger than the other two losses and dominates the training process. Once we set proper loss weights to let all tasks learn at a similar rate, as in the ``App.Opt'' baseline, both the detection and embedding tasks yield good performance. 

Second, 
results indicate that the ``Loss.Norm'' strategy outperforms the ``Uniform'' baseline but is inferior to the ``App.Opt'' baseline. The MGDA-UB algorithm, despite being the most theoretically sound method, fails in our case because it assign too large weights to the embedding loss, such that its performance is similar to the Uniform baseline. The only method that outperforms the App.Opt baseline is the weight-by-uncertainty strategy.


\begin{wrapfigure}{R}{0.5\linewidth}

    \includegraphics[width=\linewidth]{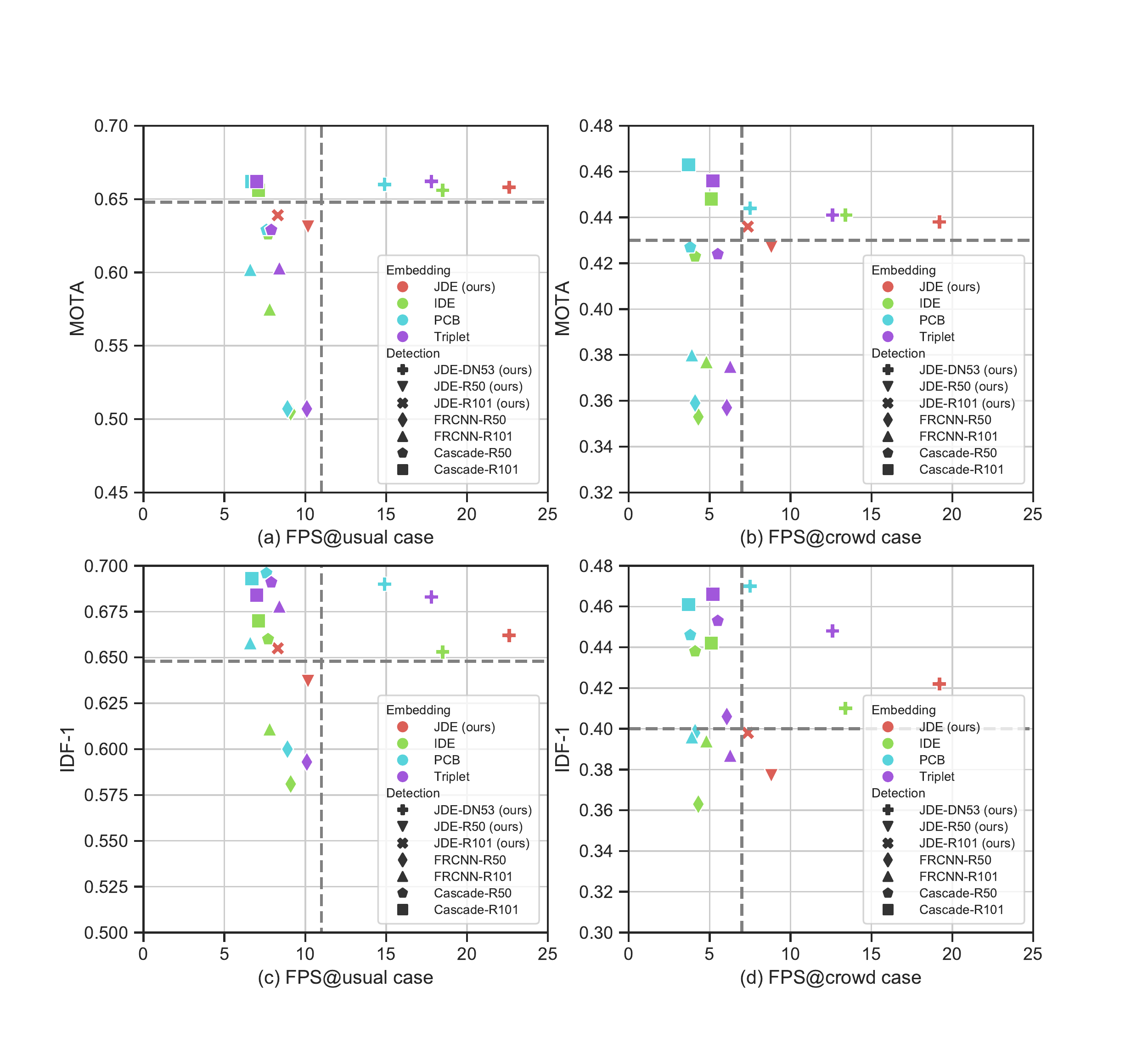}
    \caption{Comparing JDE and various SDE combinations in terms of tracking accuracy (MOTA/IDF-1) and speed (FPS). (a) and (c) show comparisons under the case where the pedestrian density is low (MOT-15 \texttt{train} set), (b) and (d) show comparisons under the crowded scenario (MOT-CVPR19-01). Different colors represent different embedding models, and different shapes denote different detectors. We clearly observe that the proposed JDE method (JDE Embedding + JDE-DN53) has the best time-accuracy trade-off. Best viewed in color.}
    \label{fig:SDE}

\end{wrapfigure}

\textbf{Comparison with SDE methods.} To demonstrate the superiority of JDE to the Separate Detection and Embedding (SDE) methods, we implemented several state-of-the-art detectors and person re-id models and compare their combinations with JDE in terms of both tracking accuracy (MOTA) and runtime (FPS). The detectors include JDE with ResNet-50 and ResNet-101~\cite{resnet} as backbone, Faster R-CNN~\cite{faster} with ResNet-50 and ResNet-101 as backbone, and Cascade R-CNN~\cite{cascade} with ResNet-50 and ResNet-101 as backbone. The person re-id models include IDE~\cite{ide}, Triplet~\cite{indefense} and PCB~\cite{pcb}. In the association step, we use the same online association approach described in Section~\ref{sec:association} for all the SDE models. For fair comparison, all the training data are the same as used in JDE. 

In Figure~\ref{fig:SDE}, we plot the MOTA metric and the IDF-1 score against the runtime for SDE combinations of the above detectors and person re-id models. Runtime of all models are tested on a single Nvidia Titan xp GPU.
Figure~\ref{fig:SDE}~(a) and (c) show comparisons on the MOT-15 \texttt{train} set, in which the pedestrian density is low, \eg, less than $20$. In contrast, Figure~\ref{fig:SDE}~(b) and (d) show comparisons on a video sequence that contains crowd in high-density (CVPR19-01 from the CVPR19 MOT challenge datast, with density $61.1$). Several observations can be made.

First, cosidering the MOTA metric, the proposed JDE produces competitive tracking accuracy meanwhile runs much faster than strong SDE combinations, reaching the best trade-off between accuracy and speed in both low-density and high-density cases. Specifically, JDE with DarkNet-53 (JDE-DN53) runs at 22 FPS and produces tracking accuracy nearly as good as the combination of the Cascade RCNN detector with ResNet-101 (Cascade-R101) + PCB embedding ($65.8\%~\vs 66.2\%$), while the latter only runs at $\sim$6 FPS. 
In the other hand, Considering the IDF-1 score which reflects the association performance, our JDE is also competitive with strong SDE combinations in the low-density case. Specifically, JDE with DarkNet-53 presents $66.2\%$ IDF-1 score at $22$ FPS, while Cascade RCNN with ResNet-101 + PCB presents $69.6\%$ IDF-1 score at $7.6$ FPS. In the high-density crowd case, performance of all methods rapidly degrades, and we observe that IDF-1 score of JDE degrades slightly more than strong SDE combinations. We find the major reason is that, in the crowd case, pedestrian often overlap with each other, and since JDE employs a single-stage detector the detected boxes often drift in such case. The misalignment of boxes brings ambiguity in the embedding, so that ID switches increase and IDF-1 score drops. Figure~\ref{fig:analysis} shows an example of such failure case.

Second, the tracking accuracy of JDE is very close to the combinations of JDE+IDE, JDE+Triplet and JDE+PCB  (see the cross markers in Figure~\ref{fig:SDE}). With other components fixed, JDE even outperforms the JDE+IDE combination. This strongly suggests the jointly learned embedding is almost as discriminative as the separately learned embedding. 

Finally, comparing the runtime of a same model between Figure~\ref{fig:SDE}~(a) and (b), it can be observed that all the SDE models suffer a significant speed drop under the crowded case. This is because the runtime of the embedding model increases with the number of detected targets. This drawback does not exist in JDE because the embedding is computed together with the detection results. As such, the runtime difference between JDE under the usual case and the crowded case is much smaller (see the red markers). In fact, the speed drop of JDE is due to the increased time in the association step, which is positively related to the target number.

\textbf{Comparison with the state-of-the-art MOT systems.} Since we train JDE using additional data instead of the MOT-16 \texttt{train} set, we compare JDE under the ``private data'' protocol of the MOT-16 benchmark. State-of-the-art online MOT methods under the private protocol are compared, including DeepSORT\_2~\cite{deepsort}, RAR16wVGG~\cite{rar}, TAP~\cite{tap}, CNNMTT~\cite{cnnmtt} and POI~\cite{poi}. All these methods employ the same detector, \ie, Faster-RCNN with VGG-16 as backbone, which is trained on a large private pedestrian detection dataset. The main differences among these methods reside in their embedding models and the association strategies. For instance, DeepSORT\_2 employs Wide Residual Network (WRN)~\cite{wrn} as the embedding model and uses the MARS~\cite{mars} dataset to train the appearance embedding. RAR16withVGG, TAP, CNNMTT and POI use Inception~\cite{inception}, Mask-RCNN~\cite{maskrcnn}, a 5-layer CNN, and QAN~\cite{qan} as their embedding models, respectively. Training data of these embedding models also differ from each other. For clear comparison, we list the number of training data for all these methods in Table~\ref{tab:sota}. Accuracy and speed metrics are also presented.

\begin{table}[t]
\centering
    \begin{tabular}{l|cc|cccccccc}
    
    \toprule
         Method & $\#$box & $\#$id &MOTA$\uparrow$ &IDF1$\uparrow$	&MT$\uparrow$ &ML $\downarrow$ &IDs $\downarrow$ &FPSD $\uparrow$&	FPSA $\uparrow$&FPS $\uparrow$ \\
         
    \midrule
         DeepSORT &	429K&	1.2k& 61.4&	62.2&	32.8&	\underline{18.2}&	\underline{781}&$<$15$^*$	&17.4&	$<$8.1 \\
         RAR16&429K&	-&	63.0&	63.8&	\underline{39.9}&	22.1&	\textbf{482}	&$<$15$^*$&	1.6	&$<$1.5 \\
         TAP&429K&	- &	64.8&	\textbf{73.5}&	\textbf{40.6}&	22.0& 	794	&$<$15$^*$ &	18.2 &	$<$8.2 \\
         CNNMTT	&	429K&	0.2K&	\underline{65.2}&	62.2&	32.4&	21.3&	946&	$<$15$^*$&	11.2&	$<$6.4 \\
         POI&	429K&	16K&	\textbf{66.1}&	\underline{65.1}&	34.0&	21.3&805&	$<$15$^*$	&9.9&	$<$6 \\
         
    \midrule
        JDE$^\texttt{864}$ &	270K&	8.7K& 62.1&	56.9&	34.4&	16.7&1,608	&\textbf{34.3}&	\textbf{259.8}&	\textbf{30.3}	 \\
        JDE$^\texttt{1088}$&	270K&	8.7K& 64.4&	55.8&	35.4&	20.0&1,544	&\underline{24.5}&	\underline{236.5}&	\underline{22.2}	 \\
    \bottomrule
    \end{tabular}
    \caption{Comparison with the state-of-the-art online MOT systems under the private data protocol on the MOT-16 benchmark. The performance is evaluated with the CLEAR metrics, and runtime is evaluated with three metrics: frames per second of the detector (FPSD), frame per second of the association step (FPSA), and frame per second of the overall system (FPS). $^*$ indicates estimated timing. We clearly observe our method has the best efficiency and a comparable accuracy.} 
    \label{tab:sota}
    
\end{table}
Considering the overall tracking accuracy, \emph{e.g.}, the MOTA metric, JDE is generally comparable. Our result is higher than DeepSORT\_2 by +3.0\% and is lower than POI by 1.7\%. 
In terms of running speed, it is not feasible to directly compare these methods because their runtimes are not all reported. Therefore, we re-implemented the VGG-16 based Faster R-CNN detector and benchmark its running speed, and then estimate the running speed upper bounds of the entire MOT system for these methods. Note that for some methods the runtime of the embedding model is not taken into account, so the speed upper bounds are far from being tight. Even with such relaxed upper bound, the proposed JDE runs at least $2\sim3\times$ faster than existing methods, reaching a near real-time speed, \ie, 22.2 FPS at an image resolution of as high as $1088\times608$. When we down-sample the input frames to a lower resolution of $864\times408$, the runtime of JDE can be further sped up to 30.3 FPS with only a minor performance drop ($\Delta$ = -2.6\% MOTA).

\begin{figure}[t]
    \centering
    \includegraphics[width=\linewidth]{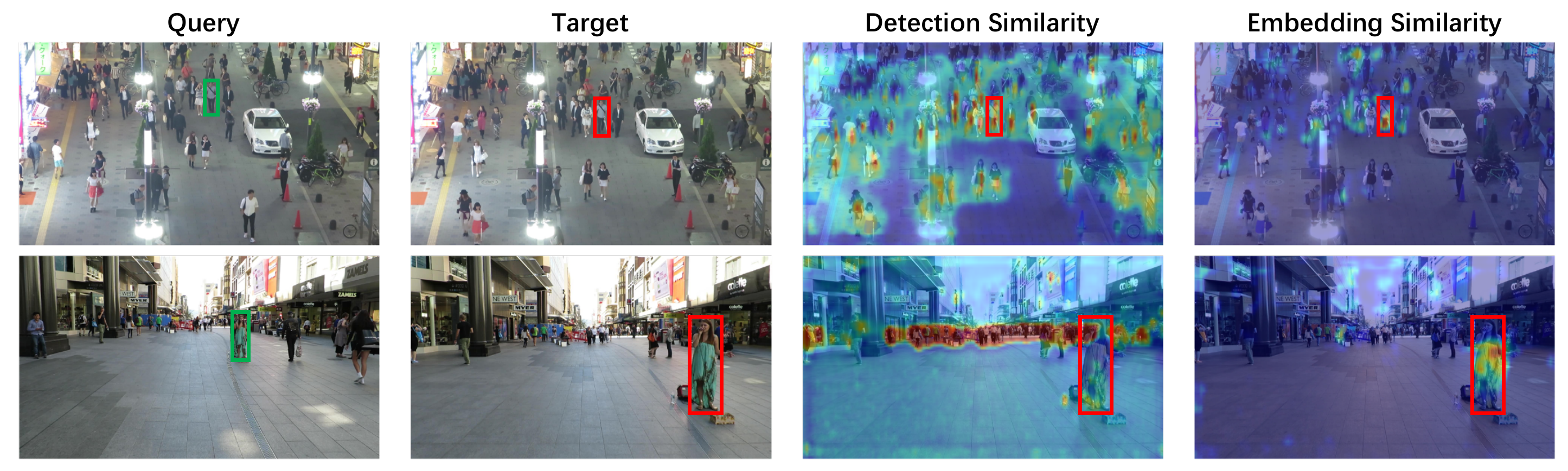}
    \caption{Visualization of the retrieval performance of the detection feature map and the dense embedding. Similarity maps are computed as the cosine similarity between the query feature and the target feature map. The joint learned dense embedding presents good correspondence between the  query and the target.}
    \label{fig:vis}
    
\end{figure}

\textbf{Visualization.} To show the discrimination of the joint learned embedding intuitively, we perform a simple retrieval experiment and visualize the results in Figure~\ref{fig:vis}. We extract the feature of a pedestrian in one video frame as a query and compute pixel-wise cosine similarity with the feature map of another frame. We compare the retrieval results between using detection feature map as the feature and using the dense embedding as the feature, and it is clearly observed the dense embedding results in better correspondence between the query and the target.

\textbf{Analysis and discussions.} One may notice that JDE has a lower IDF1 score and more ID switches than existing methods. At first we suspect the reason is that the jointly learned embedding might be weaker than a separately learned embedding. However, when we replace the jointly learned embedding with the separately learned embedding, the IDF1 score and the number of ID switches remain almost the same. Finally we find that the major reason lies in the inaccurate detection when multiple pedestrians have large overlaps with each other.  Such inaccurate boxes introduce lots of ID switches, and unfortunately, such ID switches often occur in the middle of a trajectory, hence the IDF1 score is lower. In our future work, it remains to be solved how to improve JDE to make more accurate boxes predictions when pedestrian overlaps are significant. 
    

\section{Conclusion}
In this paper, we introduce JDE, an MOT system that allows target detection and appearance features to be learned in a shared model. Our design significantly reduces the runtime of an MOT system, making it possible to run at a (near) real-time speed. Meanwhile, the tracking accuracy of our system is comparable with the state-of-the-art online MOT methods. Moreover, we have provided thorough analysis, discussions and experiments about good practices and insights in building such a joint learning framework. In the future, we will investigate deeper into the time-accuracy trade-off issue.

\clearpage
%
%
\bibliographystyle{splncs04}
\bibliography{egbib}
\end{document}